\documentclass[journal]{IEEEtran}
\usepackage{times}
\usepackage{epsfig}
\usepackage{graphicx}
\usepackage{amsmath}
\usepackage{amssymb}
\usepackage{multirow}
\usepackage{multicol}
\usepackage{booktabs}
\usepackage{soul}
\usepackage{url}
\usepackage{ulem}
\usepackage{physics}

\usepackage[pagebackref=true,breaklinks=true,letterpaper=true,colorlinks,bookmarks=false]{hyperref}

\ifCLASSINFOpdf
\else
\fi
\begin{document}
%
\title{NeRRF: 3D Reconstruction and View Synthesis for Transparent and Specular Objects with Neural Refractive-Reflective Fields}
%
%
%

\author{Xiaoxue~Chen*$^{1}$, 
        Junchen~Liu*$^{2}$,
        Hao Zhao$^{1}$,
        Guyue Zhou$^{1}$,
        and Ya-Qin Zhang$^{1}$

        \thanks{$^{1}$ Institute for AI Industry Research (AIR), Tsinghua University, China chenxx21@mails.tsinghua.edu.cn, 
        zhaohao@air.tsinghua.edu.cn.}%
\thanks{$^{2}$ Beihang University, China  
        20373790@buaa.edu.cn.}
\thanks{$^{*}$ Equal contribution.}
}

%
%

\markboth{Journal of \LaTeX\ Class Files,~Vol.~14, No.~8, August~2023} 
{Shell \MakeLowercase{\textit{et al.}}: Bare Demo of IEEEtran.cls for IEEE Journals}
%



\maketitle

\begin{abstract}
Neural radiance fields (NeRF) have revolutionized the field of image-based view synthesis. However, NeRF uses \textbf{straight rays} and fails to deal with complicated light path changes caused by refraction and reflection. This prevents NeRF from successfully synthesizing transparent or specular objects, which are ubiquitous in real-world robotics and A/VR applications. In this paper, we introduce the refractive-reflective field. Taking the object silhouette as input, we first utilize marching tetrahedra with a progressive encoding to reconstruct the geometry of non-Lambertian objects and then model refraction and reflection effects of the object in a \textbf{unified} framework using Fresnel terms. Meanwhile, to achieve efficient and effective anti-aliasing, we propose a virtual cone supersampling technique. We benchmark our method on different shapes, backgrounds and Fresnel terms on both real-world and synthetic datasets.  We also qualitatively and quantitatively benchmark the rendering results of various \textbf{editing} applications, including material editing, object replacement/insertion, and environment illumination estimation. Codes and data are publicly available at \href{https://github.com/dawning77/NeRRF}{NeRRF}.
\end{abstract}


%
\IEEEpeerreviewmaketitle

\begin{figure}[t]
\centering 
\includegraphics[width=1\linewidth]{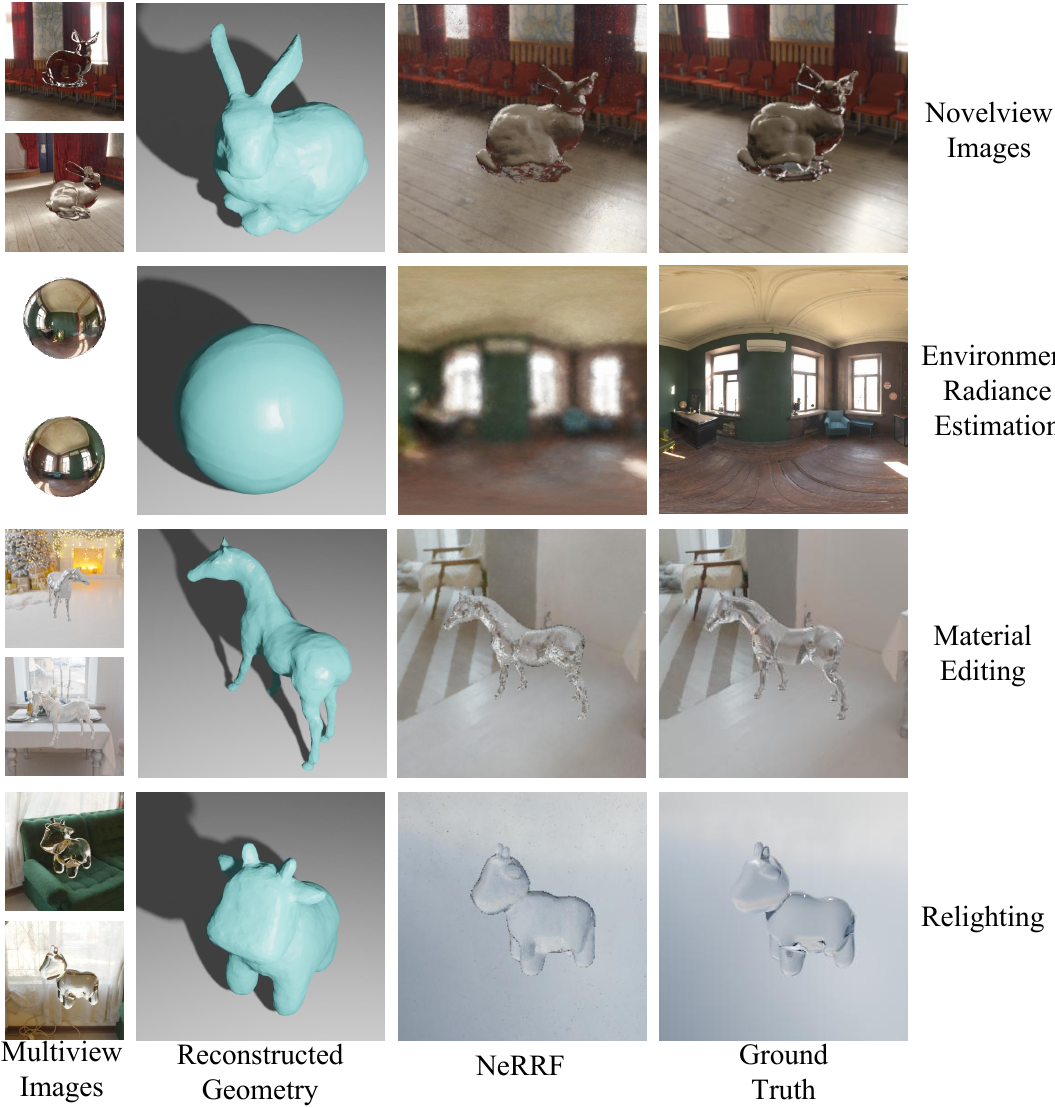}
\caption{Qualitative results of NeRRF. (a) Results on novel view synthesis. (b) The results of environment radiance estimation. We learn the radiance from only the reflection on the ball. (c) The results of material editing. We change the material of the horse from reflective to refractive. (d) The results of relighting. We relight the cow with a new illumination.}
\label{fig:teaser}
\end{figure}

\section{Introduction}
%
%
%
%

\IEEEPARstart{T}ransparent and specular objects are omnipresent in the real world, for example, glass objects, mirrors, or liquids. Successfully modeling scenes containing these objects will benefit many augmented reality or virtual reality applications including relighting or scene editing. However, 3D reconstruction and novel view synthesis on such scenes is an ill-posed problem due to complex optical phenomena, especially refraction and reflection. Different from diffuse objects that scatter incident light, the appearance of a transparent/specular object is not determined by itself but by the surrounding environment. The light bends inside (or on the surface of) the object, and the distorted environment radiance accumulated in the outgoing ray becomes the object's appearance. Therefore, the RGB information of the transparent/specular object is highly view-dependent and not reliable, and common computer vision algorithms designed for opaque objects cannot be applied to these scenes. To this end, this study aims to model scenes containing non-Lambertian objects, reconstructing the 3D geometry of these objects and learning environment radiance from reflective or refractive surfaces.

In order to model refraction and reflection and engage in physically-based ray tracing, a representation of the object's surface is necessary to be constructed, either explicitly or implicitly. Conventional 3D reconstruction techniques, such as laser scanners and standard depth sensors, are designed primarily for capturing opaque objects. Due to the intricate visual properties, the captured data of transparent objects using these sensors is distorted. Besides, as mentioned before, without known environment illumination, RGB information is also hard to utilize.  Therefore, we choose a direct and robust input to reconstruct the 3D shape of both transparent and specular objects, which is the silhouette of the object. These silhouettes can be easily obtained with on-the-shelf instance segmentors\cite{removebg}\cite{applecutout}. Besides, although it‘s difficult for humans to precisely model non-Lambertain objects from images, it's easy to distinguish the boundaries of objects, so accurate silhouettes can also be acquired through manual annotation. 

Nonetheless, how to utilize object masks is also a problem. This necessitates the underlying algorithm to have the ability to project 3D objects into mask images in a differentiable manner. Consequently, to address this requirement, we resort to the field of differentiable rendering\cite{kato2020differentiable}. NDR \cite{munkberg2022extracting},  as a recent work in this domain, employs differentiable marching tetrahedrons to enable gradient-based optimization directly on the surface mesh. As inspiration, we leverage Deep Marching Tetrahedra (DMTet) \cite{shen2021dmtet}, a differentiable hybrid shape representation, to render object mask images, which serves as the first stage. The object geometry is optimized by back-propagating the loss between the rendered masks and preprocessed silhouettes. Since there are only silhouettes as supervision, in order to avoid excessive details on the object surface, we also employ a progressive encoding that learns the object geometry in a coarse-to-fine manner.  With the above algorithm, we can successfully reconstruct the geometry of transparent/specular objects. Experiments on diverse objects (Fig. \ref{fig:geometry_comparison}) demonstrate the robustness and generalization of selecting silhouettes as input. With the reconstructed object's shape, we are enabled to disentangle the geometry and appearance, which facilitates a concentrated investigation into the environmental radiance estimation.

\begin{figure*}[t]
\centering 
\includegraphics[width=1\linewidth]{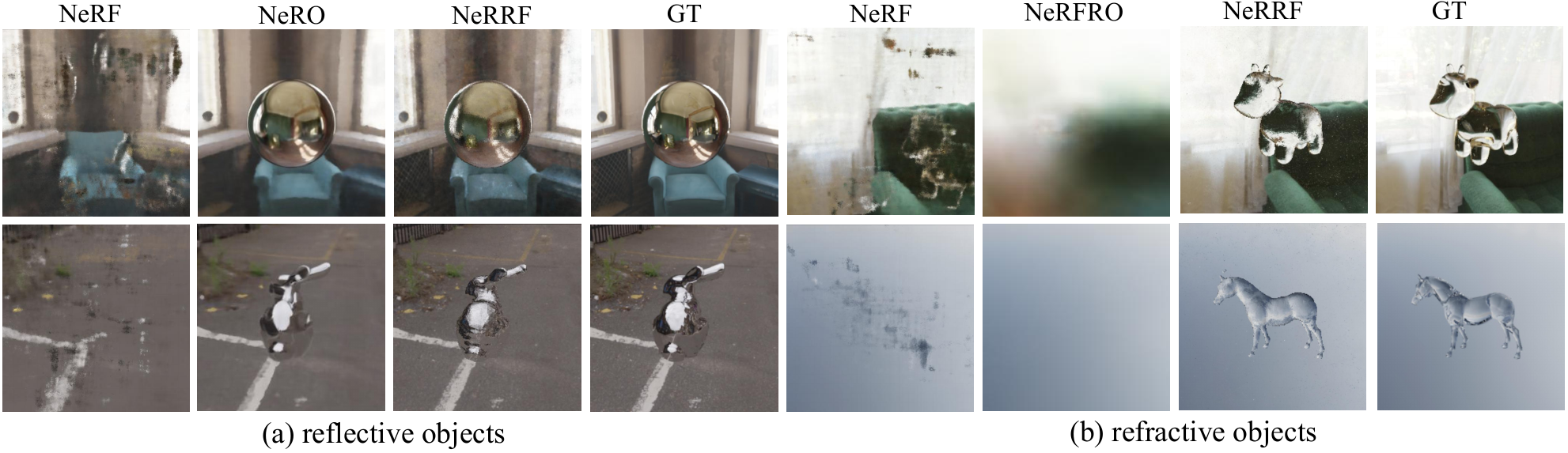}
\caption{Ground-truth and predicted novel view of all objects. The reflective objects are compared with NeRF\cite{mildenhall2021nerf} and NeRO\cite{liu2023nero} while the refractive objects are compared with NeRF\cite{mildenhall2021nerf} and NeRFRO\cite{pan2022sampling}.}
\label{fig:NVS}
\end{figure*}

Radiance estimation is a long-existing computer vision problem and neural radiance fields (NeRFs)\cite{mildenhall2021nerf} have recently broken its bottleneck. Despite the noticeable recent progress, current NeRF formulations are still limited to sampling points on straight rays for training and inference. This limitation prevents NeRF from modeling scenes containing transparent/specular objects, as these objects inevitably lead to complicated light path changes due to refraction or reflection. This limitation is evidenced by the poor results in the left column of Fig.\ref{fig:NVS} (a) and (b). Nonetheless, given that the object's shape has been estimated during the first stage, we can explicitly model the light path change in a physically-based manner, and equip NeRF with the capability of modeling \textbf{non-straight rays}, which serves as the second stage. 

Our solution, which successfully synthesizes unseen views of transparent/specular objects as shown in Fig. \ref{fig:teaser}, is named as \textbf{Neural Refractive-Reflective Field} (NeRRF). NeRRF explicitly incorporates the reflection equation and Snell’s law, the image formulation for transparent objects, into the ray-tracing process of NeRF. Due to the optical law that the relative fraction of reflection is determined by Fresnel's equation, refraction and reflection of NeRRF are modeled in a \textbf{unified} manner based on the Fresnel term. Since both refraction and reflection are highly sensitive to the surface normal, and the normal extracted from the estimated geometry in the first stage may be inaccurate, we also propose a supersampling module based upon a \textbf{virtual cone} idea. This module addresses the high-frequency noise raised by the unsmooth estimated normals. Experiments demonstrate that this module brings better results while being tractable.

Through the two-stage pipeline, NeRRF demonstrates its capability to effectively disentangle a transparent/specular object's geometry and appearance. This distinctive feature equips NeRRF to accommodate a range of AR or VR applications. As shown in Tab. \ref{tab:Method Comparison}, to the best of our knowledge, NeRRF is the only methodology capable of concurrently addressing the challenges of modeling refraction and reflection while also facilitating applications such as relighting and scene editing.

To evaluate the effectiveness of NeRRF, we build and contribute a benchmark with diverse shapes, backgrounds, and Fresnel terms, to the community. Experiments (Fig.\ref{fig:NVS}) on this benchmark conclusively demonstrate the superiority of NeRRF over the vanilla NeRF and other methods that can model refraction and reflection. In addition, we show many different applications of NeRRF, three of which are highlighted in Fig.\ref{fig:teaser}. In the second row of Fig.\ref{fig:teaser}, we achieve faithful environment radiance estimation results, which can potentially serve as a light probe \cite{debevec2008rendering}. In the third row of Fig.\ref{fig:teaser}, we edit the material of the horse by switching it from reflective to refractive. Besides, in the last row of Fig.\ref{fig:teaser}, we also demonstrate the result of the relighting of a cow, which has a similar appearance compared with the reference image generated using the rendering engine. These applications are all quantitatively evaluated on our benchmark, to facilitate future research.

Our main contributions are summarized as follows:

\begin{itemize}
    \item We reconstruct the 3D shape of non-lambertian object leveraging a hybrid representation with only object silhouette as input. With the help of progressive encoding, our method can estimate the geometry of object precisely.
    \item We propose a novel refractive-reflective field that explicitly models light path changes caused by refraction and reflection, notably in a unified manner through a Fresnel term. It also benefits from an efficient and effective anti-aliasing module using virtual cones.
    \item We contribute a diverse benchmark and conclusively show the superiority of RRF over the vanilla NeRF and a strong NeRF upgrade that integrates ray tracing. 
    \item We provide several different applications covering material editing, object replacement/insertion and environment radiance estimation. They are also quantitatively evaluated. Codes and data will be released.
\end{itemize}

\section{Related Works}

\begin{table}
\caption{Method Comparison.}
\centering
\begin{tabular}{c|ccccccc}
\toprule
    Method & A & B & C & D & E & F & G\\
\midrule
    Eikonal \cite{bemana2022eikonal} & \checkmark & & & \checkmark & & &\\
    NeRFRO \cite{pan2022sampling} & \checkmark & & & \checkmark & \checkmark & & \\
    NeTO, ... \cite{li2023neto, lyu2020differentiable, wu2018full, xu2022hybrid, wang2023nemto} & \checkmark & & & \checkmark & \checkmark & \checkmark &\\
\midrule
    NeRO, ...\cite{tiwary2022orca, liu2023nero, munkberg2022extracting} & & \checkmark & \checkmark & \checkmark & \checkmark & \checkmark &\\
     Ref-NeRF, ... \cite{boss2021nerd, verbin2022ref, srinivasan2021nerv} & & \checkmark & \checkmark & \checkmark & & \checkmark & \checkmark\\
    NeRF-DS, ...\cite{yan2023nerf, yin2023multi, guo2022nerfren} & & \checkmark & & \checkmark & & &\\
    IDR \cite{yariv2020multiview} & & \checkmark & & \checkmark & \checkmark & &\\
    PhySG \cite{zhang2021physg} & & \checkmark & \checkmark & \checkmark & \checkmark & \checkmark & \checkmark \\
\midrule
    \textbf{NeRRF(ours)} & \checkmark & \checkmark & \checkmark & \checkmark & \checkmark & \checkmark & \checkmark \\
\bottomrule
\end{tabular}
\vspace{1mm}

The first group focus on modeling the refractive effects and the second group focus on the reflective effects. (A) model the refractive effects. (B) model the reflective effects. (C) allow environment radiance estimation while training. (D) novel view synthesis. (E) shape reconstruction. (F) relighting. (G) scene editing.
\label{tab:Method Comparison}
\end{table}

\textbf{Novel view synthesis} is the task of synthesizing images or videos from unseen viewpoints given a set of input views. Early works\cite{chen1993view,levoy1996light,gortler1996lumigraph} address image-based rendering with view interpolation and light field techniques. Some other works resolve novel view synthesis by establishing better representations for 3D scenes like mesh-based\cite{waechter2014let,buehler2001unstructured,debevec1996modeling} or volumetric representations\cite{kutulakos2000theory,seitz1997photorealistic}.  With the rise of deep learning, many learning-based methods \cite{kalantari2016learning,flynn2016deepstereo,flynn2019deepview,lombardi2019neural,zhou2018stereo,mildenhall2019local} emerge and synthesize photorealistic rendering by utilizing the differentiable rendering pipeline. Recently, the neural radiance field (NeRF) \cite{mildenhall2021nerf} has shown great success in novel view synthesis. It represents a 3D scene as a continuous 5D radiance field function
parameterized by a neural network. However, training NeRF models requires a large set of images, so several works \cite{yu2021pixelnerf,xu2022sinnerf,deng2022depth} are proposed to reduce the number of training views and improve generalization by utilizing image features or depth supervision. Besides, \cite{oechsle2021unisurf,wang2022neuris} combine NeRF with implicit 3D representations which enable both surface and volume rendering using the same model. In addition,  \cite{barron2021mip,barron2022mip} focus on training an anti-aliasing NeRF with the integrated positional encoding of conical frustums. 

\textbf{Illumination estimation} has been a long-standing problem. \cite{debevec2008rendering} first use a mirrored sphere as a physical probe to measure the environment radiance. Subsequent works \cite{debevec2012single,weber2018learning,park2020seeing} use other known objects as light probes to capture the light. With the rising of deep learning, a series of works like \cite{song2019neural,zhan2021sparse,wang2022stylelight} estimate the environment lighting from LDR RGB images using end-to-end neural networks. Besides, \cite{chen2022text2light} generate high-quality HDRIs using a text-driven framework. Estimating the illumination can benefit many applications in augmented reality (AR) or virtual reality (VR). \cite{liu2020learning} disentangles outdoor scenes into illumination and other scene factors, and changes lighting effects to generate novel images. In addition, \cite{somanath2021hdr,zhao2020pointar} estimate an HDR environment map and show applications for augmented reality like object insertion. 

\textbf{Modeling reflection and refraction}. In computer graphics, reflection and refraction are modeled by simulating the light paths using techniques like ray-tracing. Hence existing works achieve reconstructing shapes \cite{kutulakos2005theory, ihrke2010transparent,morris2011dynamic,tanaka2016recovering,ma2014transparent,li2020through,han2015fixed,treibitz2011flat,xiong2021wild} or estimating poses \cite{haner2015absolute,hu2021absolute,cassidy2020refractive}, depths \cite{chen2011self} and optical flow \cite{yang2016robust} of reflective objects by simulating the light transport passing refractive materials. Besides, \cite{kutulakos2000theory,zuo2015interactive} leverage silhouette to reconstruct shapes of reflective or refractive objects, and many works like \cite{yang2019my,mei2020don,mei2021depth,mei2022glass} have been proposed to detect or segment non-Lambertian objects. For reflection, \cite{nishino2001determining} models a shiny object by separating the view-dependent and view-independent components of the surface reflection. Other works model reflection by employing a parametric BRDF\cite{kim2017lightweight}, or decomposing each image into a transmitted layer and a reflected layer\cite{sinha2012image,guo2014robust}.


While NeRF \cite{mildenhall2021nerf} maps 3D coordinates to corresponding radiance, it's challenging to deal with complex view-dependent effects as reflection and refraction. For refraction, \cite{lyu2020differentiable, li2023neto, wu2018full, xu2022hybrid} model the light path changing with a specific image capture setup such as a gray-coded background, to reconstruct the geometry of transparent shapes, which cannot be applied to images in the wild. \cite{pan2022sampling,bemana2022eikonal} follow the ray equation of Eikonal rendering \cite{ihrke2007eikonal} to conduct light transport in refractive objects. \cite{wang2023nemto} introduces a ray bending network to predict the refracted ray, while the environment radiance is considered as a prior. For reflection, \cite{liu2023nero, boss2021nerd, verbin2022ref, srinivasan2021nerv,munkberg2022extracting, zhang2021physg, yariv2020multiview} assume opaque surfaces with BRDF, and generate renderings of a non-Lambertian object. However, these methods lack the modeling of light-path changing that can’t be applied in complex reflection and refraction. Instead, \cite{tiwary2022orca,guo2022nerfren} physically models view direction changes on the object's surface, which enables representing mirror reflection. Besides, \cite{yin2023multi} use a group of feature fields in parallel sub-spaces to represent scenes with reflection. \cite{yan2023nerf} focus on dynamic specular objects and reformulate the neural radiance field to be conditioned on surface position and orientation in the observation space. 

It is manifest from Tab. \ref{tab:Method Comparison} that all the methods mentioned above have certain constraints when applied in practical scenarios. In light of these limitations, in this paper, we introduce NeRRF, a continuous neural reflective-refractive radiance field.  As illustrated in Tab. \ref{tab:Method Comparison}, NeRRF can not only reconstruct the geometry of non-Lambertian objects but also model the intricate phenomena of refraction and reflection simultaneously, which enables novel view synthesis and illumination estimation. Benefit from the disentanglement of geometry representation and radiance representation, our method can also be applied to a series of AR/VR applications including object editing, which needs to change the geometry or optical properties of object, or relighting, which needs to change the background.

\section{Method}

\subsection{Problem Formulation}

\begin{figure}
\centering
\includegraphics[width=1\linewidth]{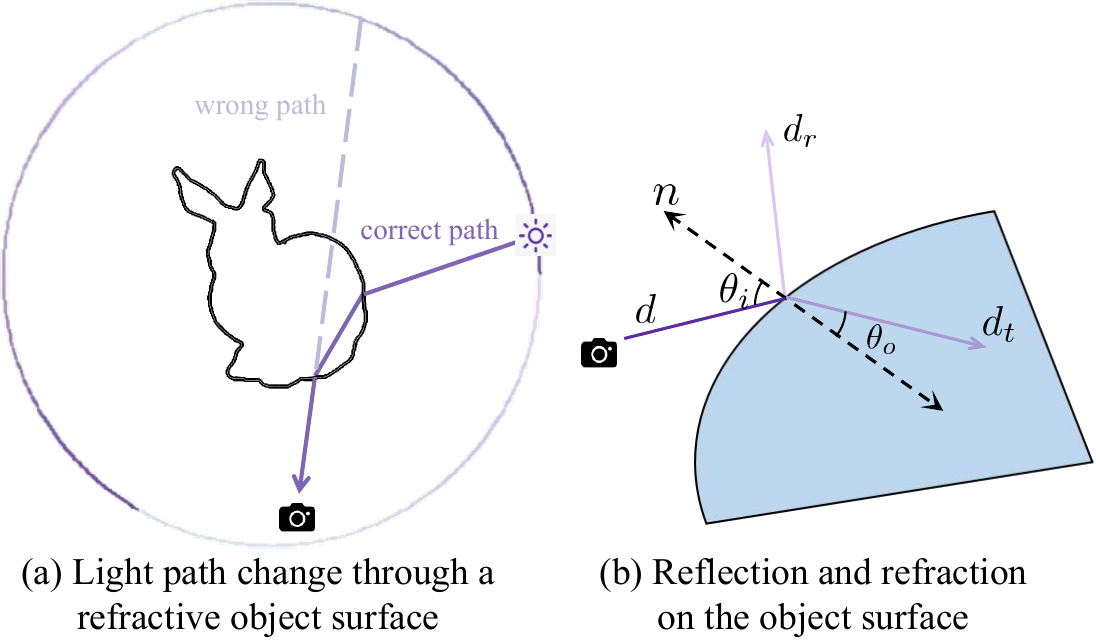}
\caption{The illustration of (a) light path change, and (b) physical process of reflection and refraction.}
\label{fig:illustration_path}
\end{figure}

 As basic optical phenomena, reflection and refraction of non-Lambertian objects are ubiquitous in the real world, and at times, such objects can serve as  a light probe \cite{debevec2008rendering} to measure the environment radiance. Given the distorted imaging on the surface of a transparent or specular object, humans can decompose the radiance based on object geometry, and infer scenes that are occluded or out-of-view.  However, for machines, handling refraction and reflection still requires exploration. 
 
In this paper, we explore how to learn environment radiance from reflective or refractive objects. Specifically, we focus on modeling a scene containing a non-Lambertian object using a physically correct representation. Given a set of posed images of the scene, we aim at reconstructing the non-Lambertian object and measuring the environment radiance,  which enables synthesizing novel view images that adhere to the laws of optical imaging, and also benefits many applications requiring exact modeling of reflection or refraction.

Recently, NeRF \cite{mildenhall2021nerf} has shown great success in novel view synthesis, which represents a 3D scene as a radiance field parameterized by a neural network $\rm f_\theta$. Specifically, given a 3D coordinate x and a viewing direction d:
\begin{align}
\rm f_\theta(x,d) = (c,\sigma),
\label{eq:nerf}
\end{align}
where $\sigma$ is a volume density and c is the corresponding RGB color value c. Despite NeRF's view-dependent radiance formulation can represent non-Lambertian effects to some extent, NeRF is completely unable to deal with complicated light path changes. Both reflection and refraction are highly sensitive to view direction. For scenes containing non-Lambertian objects, the training images are inconsistent across multiple views, hence NeRF fails in learning correct representation of the scene and the synthesized images are also physically incorrect, as shown in the middle column of Fig. \ref{fig:illustration_path} (a).

\begin{figure*}
\centering 
\includegraphics[width=1\linewidth]{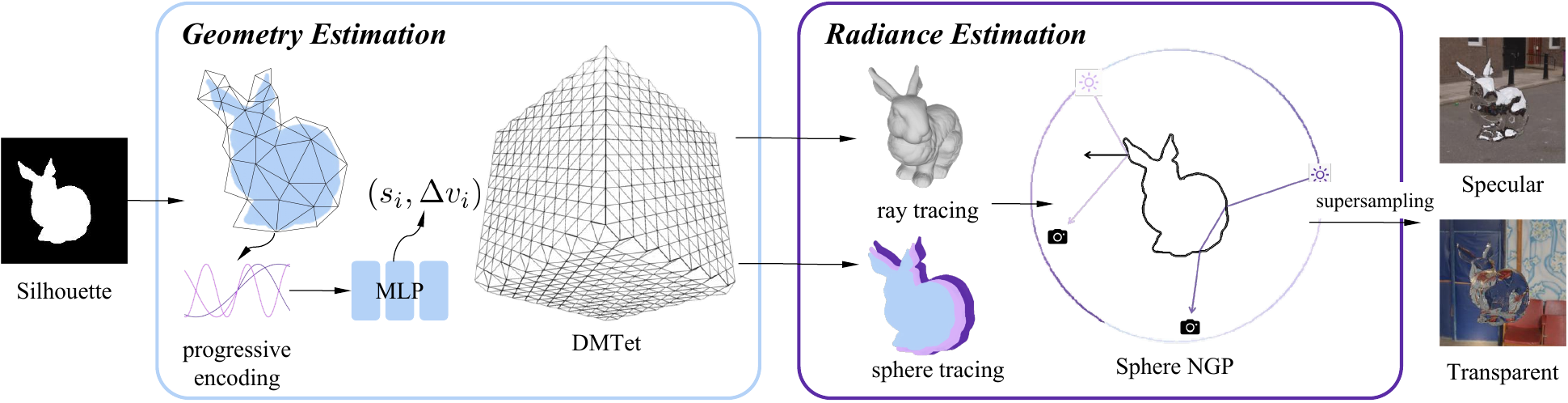}
\caption{Overall architecture. (a) Geometry estimation. With only the object mask as prior, we employ Deep Marching Tetrahedra\cite{shen2021dmtet}, a hybrid shape representation to reconstruct the geometry of the object, in which we progressively encode multilayer perceptrons (MLP) to predict the signed distance field (SDF) and per-vertex offsets defined on a grid for both smoothness and high-frequency details. (b) Radiance estimation. We propose Sphere-NGP, a direction-agnostic NGP model for radiance estimation. When a ray intersects with an object, we use the Fresnel equation to calculate the outgoing direction and radiance of the light. Supersampling is also used to remove high-frequency artifacts.}
\label{fig:main}
\end{figure*}

Therefore, we first estimate the object geometry leveraging a differentiable hybrid shape representation. Then we employ a NGP variant as the baseline to learn environment radiance and propose a physical-based ray-tracing module that models refraction and reflection in a unified manner.  With this module, we can successfully synthesize novel images by adjusting the material and geometry of the object, and recover out-of-view scenes from distorted reflection. Since the normal of object surface extracted from the estimated geometry may be inaccurate which leads to high-frequency noise in the rendering results,  we also use a supersampling strategy to address aliasing. Fig. \ref{fig:main} presents an overview of our method.

\subsection{Shape Reconstruction of Non-Lambertian Object }

Both refractions and reflections are highly sensitive to object geometry, particularly the surface normals. Thus having an accurate object shape is crucial for image synthesis. However, acquiring the geometry of non-Lambertian objects is quite challenging. Unlike opaque objects, estimating the normals and depths of non-Lambertian objects poses formidable challenges. Traditional 3D sensors, such as RGB-D cameras, are rendered ineffective due to the intricate visual properties on the object's surface, particularly the Fresnel effect. This optical phenomenon leads to distortions in the captured data, rendering it unfeasible to obtain accurate depth and normal maps from materials influenced by this effect. Meanwhile, the appearance of a transparent/specular object is also hard to model without known environment radiance due to the view-dependent effects.

Instead of estimating geometry from view-dependent RGB images or distorted depths, we directly recover the object shape from its multi-view silhouette, which is the most robust information extracted from non-Lambertian objects and can be easily obtained using on-the-shelf instance segmentors\cite{removebg} as well.

\subsubsection{Shape Representation}
To learn the object geometry solely from the object mask, we require a differentiable rendering pipeline capable of rendering the geometry's silhouette. This allows us to back-propagate the loss and optimize the object geometry using only the object mask loss. To do so, we integrate Deep Marching Tetrahedra\cite{shen2021dmtet} (DMTet), a differentiable hybrid surface representation in our geometry estimation pipeline. DMTet represents geometry as a signed distance field (SDF) defined on a deformable tetrahedral grid $T$. Each vertices $v_i$ in the grid $T$ are characterized by both the SDF value $S_i \in R$ and the deformation $\Delta v_i \in R^3$. The grid is deformed as $v_i' = v_i + \Delta v_i$ and can leverage all its resolution. Using the differentiable marching tetrahedra, explicit mesh can be extracted by computing the SDF level set conditioned in each deformed tetrahedron $T_k \in T$.

Benefited from the differentiable representation, we can render object mask images and back-propagate the loss to optimize the object geometry characterized in the per-vertex SDF value $s_i$ and deformation value $\Delta v_i$. Given the estimated binary object mask $\mathbf{\hat{M}}$ and the ground truth mask $\mathbf{M}$, the mask loss is formulated as: 

\begin{align}
\mathcal{L}_{\text {mask}}= \left\| \mathbf{M} - \mathbf{\hat{M}} \right\|_2
\label{eq:mask_loss}
\end{align}
 
\subsubsection{Progressive Encoding}
Despite the advantages, directly optimizing the per-vertex SDF values $S_i$ and deformations $\Delta v_i$ conditioned in parameter grid can introduce high-frequency noise in the surface mesh (refer to Sec. \ref{section:pe}). Compared with the per-vertex parameterized method used in \cite{munkberg2022extracting}, we employ a MLP with fewer frequency encoding to predict the SDF and deformation $(s_i, \Delta v_i)$ of each tetrahedral vertex can reduce the surface high-frequency noise. However, reducing the surface high-frequency noise may result in the loss of important high-frequency surface details, which are critical for achieving high-quality surface reconstruction. 

To strike a balance between reducing high-frequency noise and preserving shape details, we leverage a progressive encoding strategy inspired by SAPE \cite{hertz2021sape} , which progressively unveils higher frequency encoding channels as the training iterations increase. With an input position $p$ and a set of encoding functions $\left\{\mathbf{e}_n\right\}_{n=1}^N$, we applied a set of soft mask $\left\{\mathbf{\alpha}_n\right\}_{n=1}^N$ to each encoding. We use the following rule to config $\alpha_n$ at the time step $t$:
\begin{align}
\alpha_n(t)= \begin{cases}1 & n \leq n_{\text {base }} \\ clamp(\frac{t-t_s(n)}{t_e(n)-t_s(n)}, 0, 1) & n>n_{\text {base }}\end{cases}
\label{eq:progressive_mask}
\end{align}

The lowest $n_{base}$ encoding channels are always exposed, while the remaining encoding channels will be gradually revealed during $[t_s(n), t_e(n)]$. The values of $t_s(n)$ and $t_e(n)$ increase with the frequency $n$. Benefiting from the progressive encoding, our model is capable of learning the lower frequency features at earlier stages and higher frequency details later on, which significantly reduces the high-frequency noises.

\subsubsection{Mesh Extraction} 
Benefit from the differentiable rendering pipeline, the implicit SDF based representation can be converted into explicit mesh using the marching tetrahedra algorithm\cite{shen2021dmtet}. Specifically, with vertex SDF value $s_i$ and deformed vertex position $v_i '$, the surface can be detected based on the signs of $s_i$. For an edge $e$ with 2 endpoints $v_i$ and $v_j$, if $sign(s_i) \neq sign(s_j)$, the position with zero SDF value can be calculated using linear interpolation: $\mathbf{v}_0=\frac{\mathbf{v}_i^{\prime} s_j-\mathbf{v}_j^{\prime} s_i}{s_j-s_i}$. For each point $v_0 \in V_0$ with the interpolated zero SDF value, we also applied IGR regularization to these points:

\begin{align}
\mathcal{L}_{\mathrm{eikonal}}=\mathbb{E}_{v_0  \in V_0}\left(\left\|\nabla f_\theta(v_0)\right\|_2-1\right)^2.
\label{eq:eikonal_loss}
\end{align}

The importance of the Eikonal loss in regularizing the SDF field has been demonstrated in previous studies, as shown in \cite{icml2020_2086}. When using the grid-based SDF values and deformations $(s_i, \Delta v_i)$, it is possible to calculate the SDF value of surface points through interpolation. However, this approach hinders the calculation of gradients for these surface points, thus preventing the application of the Eikonal loss. As a result, high-frequency noise on the object's surface can be observed, as illustrated in Fig. \ref{fig:MLP_based_DMTet_ablation}. Indeed, it's a crucial factor contributing to the excellent performance of our geometry reconstruction method.

The overall loss function in this stage is defined as the weighted sum of individual loss with $\lambda$ representing the weight terms:

\begin{align}
\mathcal{L}=\lambda_{\text {mask }} \mathcal{L}_{\text {mask}}+\lambda_{\text {eikonal}} \mathcal{L}_{\text {eikonal}}.
\label{eq:geometry_stage_loss}
\end{align}

\subsection{Illumination Estimation with Refractive-Reflective Field}

With the estimated shape in the first stage, we can successfully disentangle of object's geometry and appearance. Based on this, in this section, we focus on measuring the environment radiance from reflection/refraction with a neural radiance field. 

Both synthesised and real world datasets are used in the experiment, while the synthesised images are generated by a non-Lambertian object and an HDRI lighting environment through Blender \cite{Blender}. Following many works that use environment maps as the illumination representation\cite{blinn1976texture,li2020through}, we assume that the environment is infinitely far away from the object, so the radiance of the object's surface only depends on the direction $d$ of the emergent ray in this scenario.

Targeted at estimating the environment map, we propose a grid-based representation named \textbf{Sphere-NGP}, which is a variation of Instant-NGP \cite{muller2022instant}. Given a ray $r(t)=\mathbf{o}+t\mathbf{d}$, Sphere-NGP takes only the view direction as input and encodes the direction to a trainable grid feature, then the feature is mapped to its radiance using an MLP $\rm f_\theta$, which is formulated as:

\begin{align}
\rm \mathbf{c}(r) = f_\theta(e(\mathbf{d})). 
\label{eq:env}
\end{align}

Where $e(\cdot)$ is the multi-resolution hash encoding. Leveraging the advantages of the multi-resolution hash map, Sphere-NGP demonstrates the ability to render more high-frequency details within a shorter time. For additional benchmark results, refer to Sec. \ref{section: spherengp} .

For each pixel in the image, the incident ray $r(t) = \mathbf{o} + t\mathbf{d}$ passing through that pixel can be derived through camera calibration, where $\mathbf{o}$ is the camera origin and $\mathbf{d}$ is the corresponding view direction.  Certain rays will intersect with the object's surface, causing them to undergo light path deflections. Fig. \ref{fig:illustration_path} (a) depicts the alterations in the light path when a ray traverses a refractive object, in which employing the incident ray directly could result in incorrect estimation of the environment radiance.  Instead, we model reflection and refraction of rays through a physically-based ray-tracing module, and the proposed field which has a ray-tracing module is named as the \textbf{Neural Refractive-Reflective Field (NeRRF)}.

Given the geometry and material of a non-Lambertian object and the incident ray r(t), we first calculate their closest intersection point $\mathbf{p}$ and its corresponding normal $\mathbf{n}$. The estimated objects surface can be represented by either an mesh extracted from DMTet or by configuring the SDF level set through an implicit neural network $f_\theta(\mathbf{p})$. For the explicit representation, we employ ray tracing based on ray-mesh intersection to determine if the ray hits the mesh surface. And for the SDF-based representation, sphere tracing is applied to detect the intersection points alternatively. 

The normal of intersection point is also crucial for confirming the outgoing direction. In the case of the mesh representation, the intersection point's normal is interpolated from the intersecting faces for smoothness. When employing the SDF representation with an MLP, the surface normal can be obtained through its first-order derivative:  $\mathbf{n} = \nabla_p f_\theta$.

With the intersection point $\mathbf{p}$ and corresponding normal $\mathbf{n}$, the outgoing ray direction $\mathbf{d_t}$ on the surface of transparent object is determined by the Snell's law:
\begin{align}
\rm cos\theta_o  = \sqrt{1-\lambda^2(1-cos^2\theta_i) },\\
\rm \mathbf{d_t} = -\lambda  \mathbf{d} + (\lambda( \mathbf{d} \cdot \mathbf{n}) - cos\theta_o ) \mathbf{n},
\label{eq:snell}
\end{align}
where $\lambda$ is the index of reflection, $\theta_i$and $\theta_o$ are incident angle and emergent angle respectively. The direction of reflected ray $\rm \mathbf{d_r}$ is formulated as:
\begin{align}
\rm \mathbf{d_r} =  \mathbf{d} + 2cos\theta_i \mathbf{n}. 
\label{eq:refr_dir}
\end{align}

For rays that hit the surface, we replace the incident ray $r(t) = \mathbf{o} + t\mathbf{d}$ with either the reflected ray $\rm r_r(t) = \mathbf{p} +t \mathbf{d_r}$ or the refracted ray $\rm r_t(t) = \mathbf{p} +t \mathbf{d_t}$ at the intersection points, contingent upon the object's material. Then we proceed to trace the new ray until there are no new intersection points with the object. In practice, constrained by GPU memory limitations, we consider light paths with a maximum of two bounces for simplication.

When light strikes a refractive surface, both reflection and refraction of the light may occur, causing a difference  between the radiance of  the incident ray and the emergent ray. Fresnel equations provide the ratio of the reflected radiance to the incident radiance, characterized by a Fresnel term $F$:
\begin{align}
\rm F = \frac{1}{2} (\frac{\mathbf{d_i}\cdot \mathbf{n}-\lambda \mathbf{d_t}\cdot \mathbf{n}}{\mathbf{d_i} \cdot \mathbf{n}+\lambda \mathbf{d_t}\cdot \mathbf{n}})^2 & + \rm \frac{1}{2}(\frac{\lambda \mathbf{d_i}\cdot \mathbf{n}-\mathbf{d_t}\cdot \mathbf{n}}{\lambda \mathbf{d_i}\cdot \mathbf{n}+\mathbf{d_t}\cdot \mathbf{n}})^2 \\
\rm R_r = F\cdot R_i, \quad &  \rm R_t = (1-F)\cdot R_i
\label{eq:fresnel}
\end{align}

where $\rm R_i$, $\rm R_r$, and $\rm R_t$ denote the radiance of incoming, reflected, and refracted rays respectively. We consider the Fresnel effect only with the first intersection of the refractive surface. For reflective surfaces, we set $F$ to 1 directly, adhering to the principle of energy conservation.  The ultimate radiance with the incident ray $r(t)$ intersecting with an object is:
\begin{align}
\rm c(r) = F\cdot c(r_r) + (1-F)\cdot c(r_t)
\label{eq:color}
\end{align}

 For those incident rays that do not intersect with the non-Lambertian object, we directly follow Eq. \ref{eq:env} without any further processing. The training objective in this stage is formulated as:
\begin{align}
\rm \mathcal{L}_{RGB} = \frac{1}{|R|}\sum_{r \in R}\Vert c(r)-\hat{c}(r)  \Vert_2^2
\label{eq:rgb_loss}
\end{align}

where $R$ denotes a set of rays, and $\rm \hat{c}(r)$ is the ground-truth RGB value of ray $r$.

\textbf{Point-based refractive-reflective fields}. The radiance representation introduced by NeRRF,  as expressed in Eq. \ref{eq:env} is compatible with the dataset setting that uses an environment map as the illumination representation. In some cases of real-world dataset where the environment cannot be treated as infinitely distant, a radiance representation incorporating point-based considerations is necessary. Hence we introduce a variant NeRRF named NeRRF-P(oint), which adopts the point-based radiance field based on Instant-NGP \cite{muller2022instant}. The radiance formulation of NeRRF-P can be expressed as:
\begin{align}
\rm \mathbf{c}(r) = f_\theta(e(\mathbf{x}), \mathbf{d})).
\label{eq:ngp}
\end{align}
where x is the position  of the sampling points on the ray, and $\mathbf{d}$ is the viewing direction. NeRRF-P also incorporates the ray-tracing/sphere-tracing module used in NeRRF, for dealing with the real world scenario. 

\subsection{Anti-aliasing with Supersampling}

\begin{figure}
\centering
\includegraphics[width=0.6\linewidth]{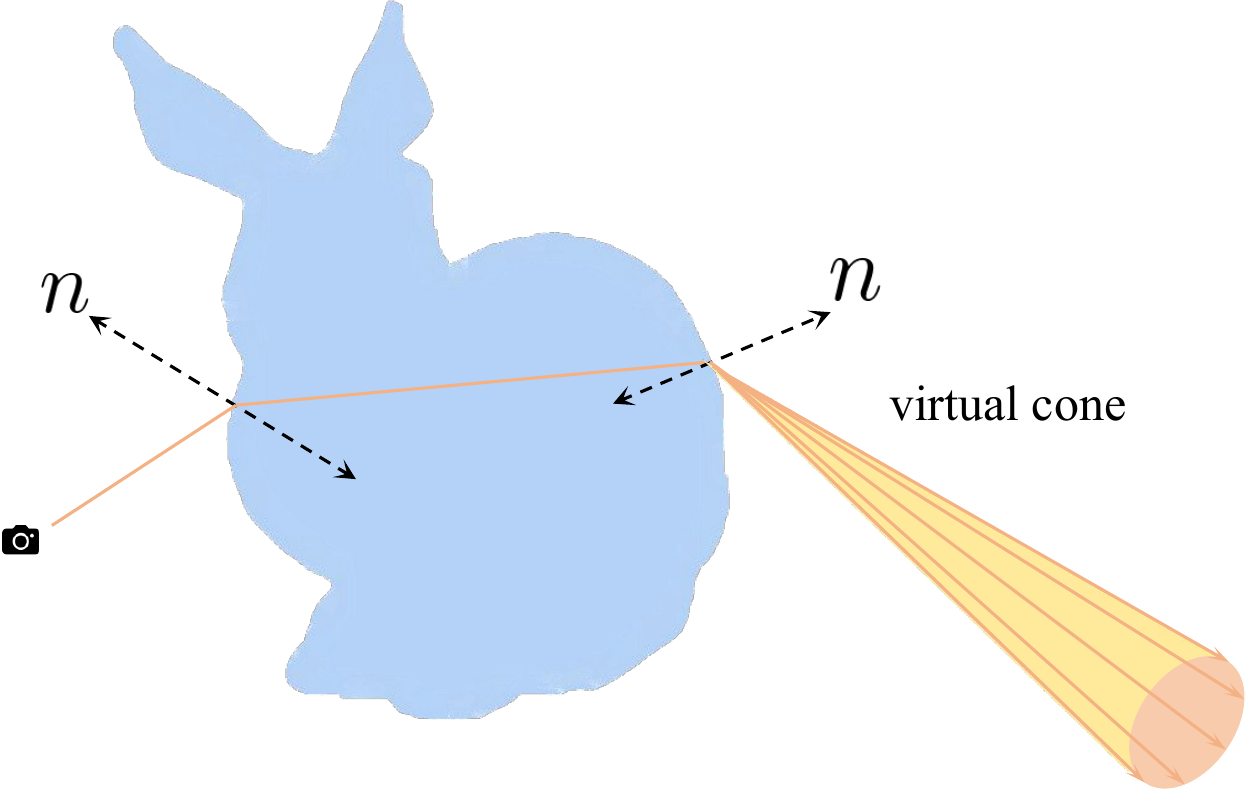}
\caption{Virtual cone supersampling illustration. After the ray is emitted from the object's surface, we employ a virtual cone for supersampling, enabling multiple sampling and anti-aliasing.}
\label{fig:illustration_supersampling}
\end{figure}

Given that the geometry estimation stage lacks direct normal supervision or a consistency constraint, focusing only on minimizing the mask loss as depicted in Eq. \ref{eq:mask_loss}, the resulting reconstructed surface normals are not as accurate. This deficiency in accuracy translates to a lack of local coherence in the normals, consequently introducing undesirable high-frequency noise onto the surface. To reduce surface high-frequency artifacts, we model a ray as a light cone when the ray is shot out of the object's surface, as illustrated in Fig. \ref{fig:illustration_supersampling}. The radiance of the object's surface then becomes the integration of all incident radiance in the virtual cone,  which can also reduce the aliasing. For the rays that do not intersect with the object, their accuracy is not affected by the inaccurate surface normals, for which we keep using Eq. \ref{eq:env} as the radiance formulation.

The axis of the virtual cone is the outgoing direction $\mathbf{d}$. In the cone, we randomly sample $N$ directions on the uniform distribution of angle and obtain the corresponding sampled radiance as Eq. \ref{eq:env}. To penalize rays that deviate significantly from the axis, we introduce a confidence index $conf$ to quantify the reliability of each ray. Within the cone, we assume that the angle between sample direction and the cone axis adhere to a Gaussian distribution $N(0, \sigma)$, with $\sigma$ being a hyperparameter. The confidence index $conf$ is defined as the probability density function (pdf) value of the Gaussian distribution corresponding to the given deviation angle. The ultimate radiance of outgoing rays $r_o$ is then represented as the weighted sum of radiance values for each ray, with the weight determined by the corresponding confidence index $conf$. The final radiance of the outgoing ray $\rm r'$(t)  is formulated as: 
\begin{align}
\rm c(r') = \frac{\sum_{i=1}^{N_s}conf_i \cdot f_\theta(\gamma(\mathbf{d_i}))}{\sum_{i=1}^{N_s}conf_i},
\label{eq:sampling}
\end{align}

$\rm \mathbf{d_i}$ is the sampled directions in the virtual cone. It is noteworthy that employing supersampling on the vanilla NeRF leads to increased complexity, while the complexity of our ray sampling strategy is the same as that of the vanilla NeRF, since we do not sample points.

\section{Experiment}

\subsection{Experiment protocol}

\subsubsection{Datasets} 
\begin{figure}[t]
\centering
\includegraphics[width=1\linewidth]{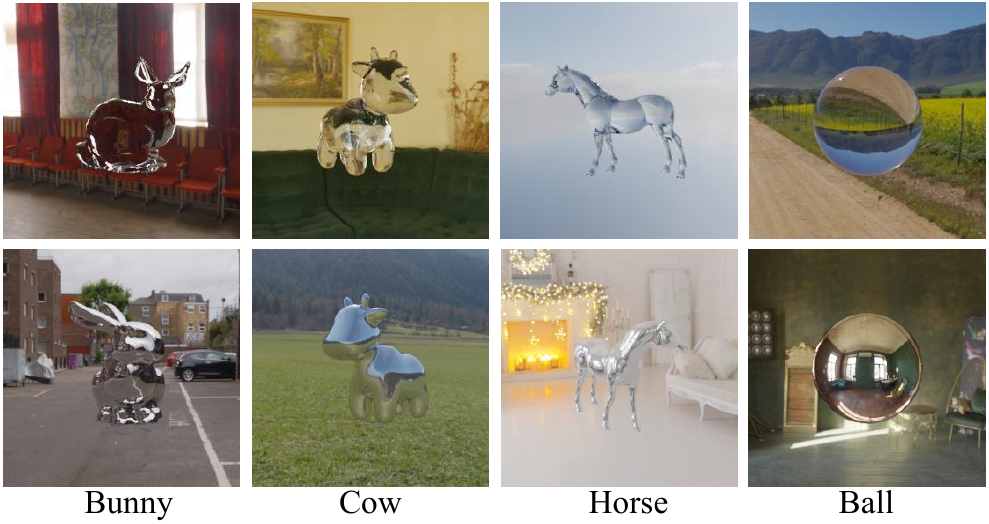}
\caption{Illustration of our Blender Synthetic Dataset.}
\label{fig:dataset}
\end{figure}

Our synthetic dataset is generated using Blender's physics engine \cite{Blender} and physically-based ray-tracing Blender Cycles rendering engine, which is capable of simulating the physically accurate non-Lambertian effects exhibited by objects, including refraction, reflection, and Fresnel effects. We use 4 objects including cow, horse, bunny and ball. Each of these objects is utilized twice, once to represent specular properties and once for transparent properties. These objects are positioned within eight scenes derived from PolyHaven\cite{polyhaven}. The index of refraction (IOR) for reflective materials is set within the range of 1.1 to 1.5. The generated images are illustrated in Fig. \ref{fig:dataset}.

We conducted sampling of 89 camera poses on the upper hemisphere around the object. For each sampled camera pose, we gathered ground truth data that encompassed: (1) monocular RGB rendering, (2) segmentation mask of the non-Lambertian object, (3) the camera's pose, and (4) object material details, including whether the object is refractive or reflective and its corresponding index of refraction (IoR). 

\subsubsection{Evaluation Details} We conducted a comprehensive benchmark of our approach on the synthesized Blender dataset, utilizing a subset of the generated data for both training and testing. Specifically, we employed 50 images for training and 39 images for testing within each scene. During the geometry estimation stage, we assessed reconstruction quality using the Chamfer Distance (CD) between the predicted surface point cloud and the ground truth surface point cloud. For the environment estimation task, we employed standard image quality metrics, including PSNR, SSIM\cite{wang2004image}, and LPIPS\cite{zhang2018perceptual} for all quantitative assessments to gauge the performance.

\subsubsection{Baselines} We compare our method with several state-of-the-arts surface reconstruction methods, including IDR \cite{yariv2020multiview}, PhySG \cite{zhang2021physg}, NDR \cite{munkberg2022extracting}, and NeRO \cite{liu2023nero} for geometry estimation task. For the novel view synthesis task, we employ NeRF \cite{mildenhall2021nerf} and NeRO \cite{liu2023nero} as our baseline methods.  Note that we did not compare with some previous arts working on transparent objects mentioned in related works, mainly because they are not open-source \cite{wang2023nemto} or have significant differences in experiment settings \cite{li2023neto, xu2022hybrid} (requiring specific backgrounds, etc.).

\begin{figure*}[t]
\centering
\includegraphics[width=0.8\linewidth]{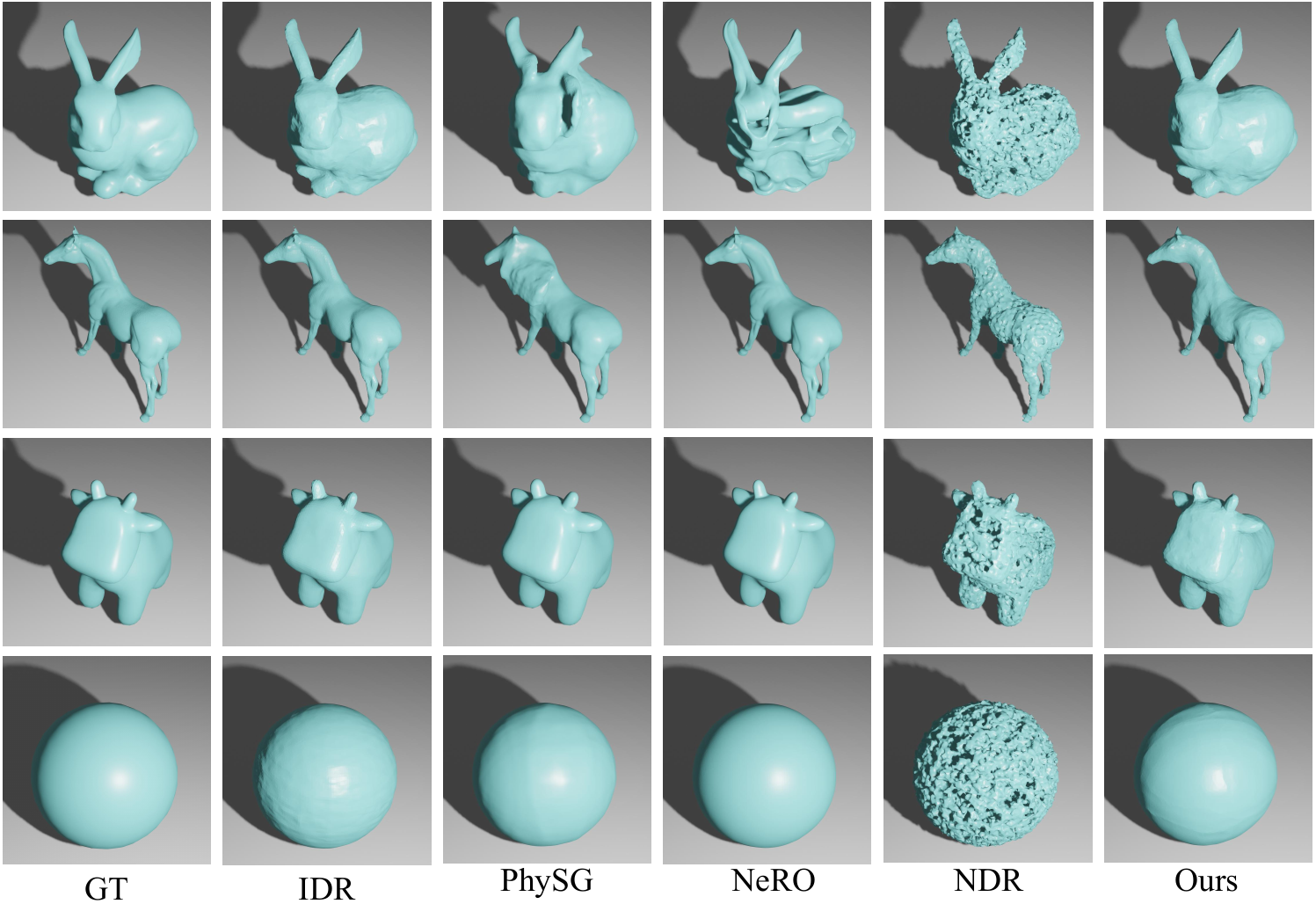}
\caption{Ground-truth and reconstructed surfaces of the Blender synthetic dataset. Our results are compared with IDR\cite{yariv2020multiview}, PhySG\cite{zhang2021physg}, NeRO\cite{liu2023nero}, and NDR\cite{munkberg2022extracting}. Note that NeRO and NDR are only capable of reconstructing glossy surfaces, so we only include reflective objects for the two methods. }
\label{fig:geometry_comparison}
\end{figure*}

\subsection{Implementation and traing details}
 We conducted our implementation using the PyTorch\cite{paszke2019pytorch} framework and employed a single NVIDIA GeForce RTX 3090 GPU for all experiments.

\begin{figure}[t]
\centering 
\includegraphics[width=1\linewidth]{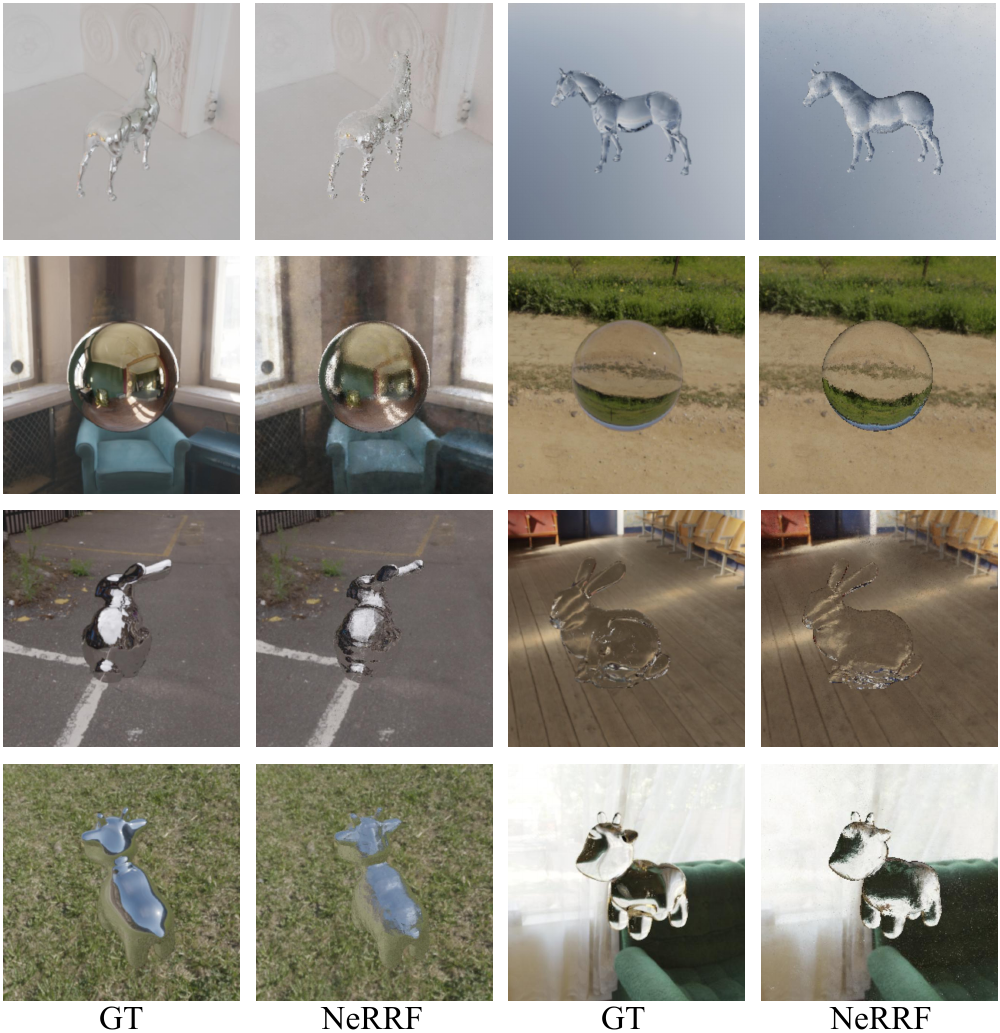}
\caption{More qualitative results of novel view synthesis.}
\label{fig:more_results}
\end{figure}
 
\subsubsection{Geometry estimation details}
In the stage of geometry estimation, we trained each object with the progressive encoding applied. We applied Fourier frequency encoding with 6 encoding channels, the input $x$ and the lowest encoding channel comprising $sin(x), cos(x)$ were constantly enabled. The remaining channels were progressively activated in 500 iteration following the Eq. \ref{eq:progressive_mask}. We employed the Adam optimizer with the learning rate of 0.001 to train the objects. We trained each objects for 3 epochs, with each epoch comprising 3000 iterations. For objects with more high-frequency details like the bunny, cow, and horse, we set the eikonal loss coefficient ($\lambda_{eikonal}$) to 0.01 for the first two epochs and 0.001 for the subsequent 3000 iterations. In the case of the ball, which exhibits fewer high-frequency details, $\lambda_{eikonal}$ was set to 0.1 for the first two epochs and 0.01 for the remaining epoch. For the baseline methods\cite{liu2023nero}, \cite{zhang2021physg} and \cite{yariv2020multiview} and \cite{munkberg2022extracting}, we adhered to their original training configurations.

\subsubsection{Radiance estimation details}
In the training of Sphere-NGP, we employed the Adam optimizer with a learning rate of 0.01. Each scene underwent four epochs, and each epoch involved 25,000 iterations. For training the NeRF backbone, we used the Adam optimizer with a learning rate of 0.005. Each object went through eight epochs, with each epoch comprising 25,000 iterations. This extensive training was necessary as NeRF can be harder to converge.Regarding other methods \cite{yariv2020multiview}, \cite{zhang2021physg}, and \cite{liu2023nero}, we adopted their original settings.

Additionally, we set the parameter $\sigma$, which determines the cone apex angle used for supersampling, to different values. For objects with specular surfaces (bunny, cow, horse), the apex angle was set to 20 degrees, while for the specular ball, it was set to 10 degrees. For transparent objects (bunny, cow, horse), the value was set to 10 degrees, and for the transparent ball, it was 5 degrees.

Our framework supports both SDF-based sphere tracing and mesh-based ray tracing. Since the mesh is extracted from DMTet during the geometry estimation stage, there is minimal difference in rendering results between the two methods. Notably, SDF-based sphere tracing offers significantly faster rendering performance.

\subsection{Geometry Reconstruction} 
\begin{figure}[t]
\centering
\includegraphics[width=0.5\linewidth]{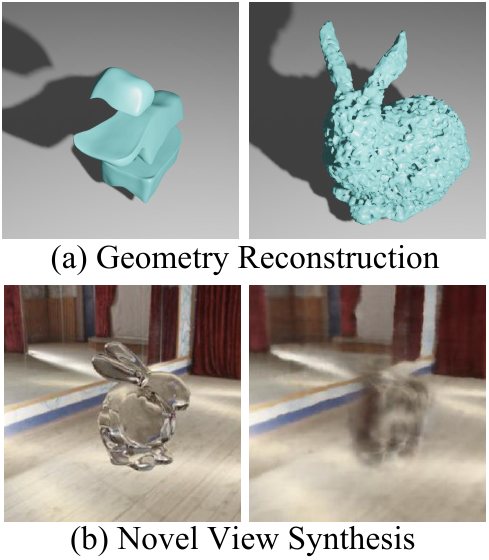}
\caption{Existing method limitation Existing method are unable to accurately reconstruct the geometry of transparent objects and generate novel view images. (a) Geometry Estimation The left side shows the reconstructed surface of a transparent cow, learned using NeRO, while the right side displays the surface of a transparent bunny, learned using NDR. (b) Novel View Synthesis The left side is the ground truth novel view image of transparent bunny while the right side shows the predicted image generated by NeRO.}
\label{fig:refr_failure_cases}
\end{figure}

We evaluate our capacity of geometry estimation for transparent/specular objects compared with previous arts, as shown in Tab. \ref{tab:geometry_comparison}. NeRRF's hybrid geometry representation demonstrates stable performance on four different objects and performs best in the average chamfer distance. This verifies our method's robustness, which is because our input only depends on the silhouettes and does not require other view-dependent inputs.   The silhouette defines the boundaries of an object, which is crucial for reconstructing the object's shape, meanwhile, it is not affected by the material and appearance of the object. This allows our method to handle objects of different materials, including reflective and refractive, or even be generalized to opaque objects, while NeRO and NDR are only capable of modeling glossy surfaces, as illustrated in Fig. \ref{fig:refr_failure_cases} (a).  In addition, Fig. \ref{fig:geometry_comparison} provides qualitative comparisons with other methods. As shown in the figure, compared to ground truth, our method can accurately reconstruct the shape of objects with enough details. Besides, NeRO can perfectly reconstruct the reflective surfaces of the horse, cow, and ball in our blender synthetic dataset. However, it fails when dealing with the bunny. This also demonstrates that the mask may be vital for the geometry estimation of some non-Lambertian objects since NeRO doesn't require object masks as input.

\begin{table}
  \caption{Comparison of geometry reconstruction methods}
  \centering
  \begin{tabular}{ccccc|c}
    \toprule
    Method & Bunny & Cow & Horse & Ball & Avg\\
    \midrule
    IDR\cite{yariv2020multiview} & \textbf{0.0032} & \underline{0.0037} & 0.0005 & \underline{0.0035} & \underline{0.0027}\\
    PhySG\cite{zhang2021physg} & 0.0209 & 0.0047 & 0.0171 & \underline{0.0035} & 0.0116 \\
    NeRO\cite{liu2023nero} & 0.0313 & \textbf{0.0024} & \textbf{0.0002} & \textbf{0.0033} & 0.0093 \\
    NDR\cite{munkberg2022extracting} & 0.0665 & 0.0555 & 0.0151 & 0.2339 & 0.0928 \\
    NeRRF(ours) & \underline{0.0038} & \textbf{0.0024} & \underline{0.0003} & 0.0037 & \textbf{0.0025}\\
    \bottomrule
  \end{tabular}
  \label{tab:geometry_comparison}
\vspace{1mm}

We report the chamfer distance on ground truth mesh and extracted meshes on the blender synthetic dataset as a quantitative measure for reconstructed geometry quality.
\end{table}

\subsection{Novel View Synthesis} 
\begin{figure}[t]
\centering 
\includegraphics[width=1\linewidth]{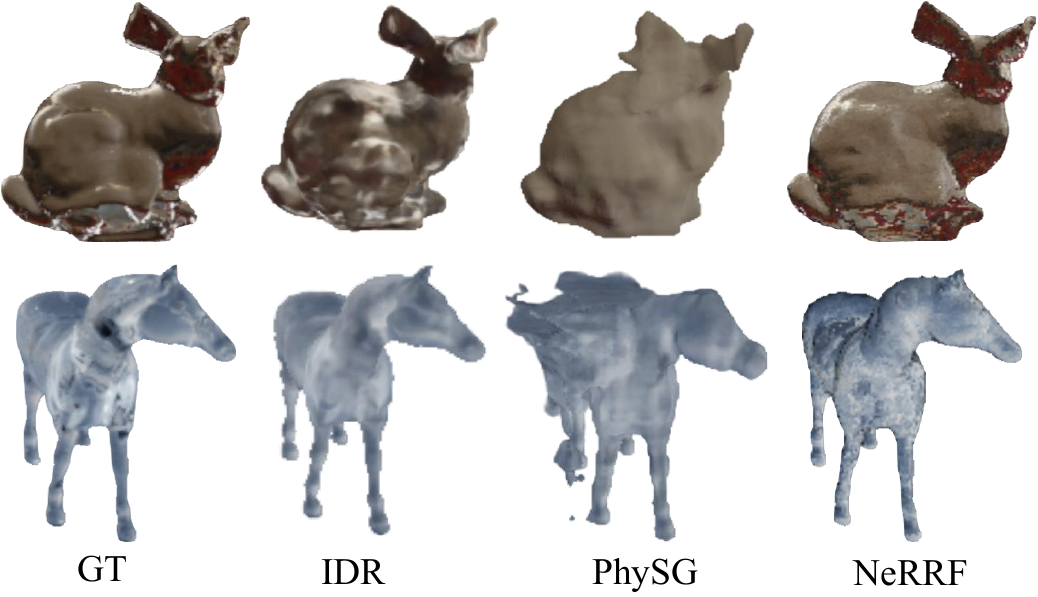}
\caption{Ground-truth and predicted novel view of refractive objects. Our results are compared with IDR\cite{yariv2020multiview} and PhySG\cite{zhang2021physg}. }
\label{fig:NVS_refractive}
\end{figure}

As IDR and PhySG do not inherently model the background, we employed them for a qualitative evaluation of novel view synthesis without including the background. While IDR\cite{yariv2020multiview} and PhySG\cite{zhang2021physg} can estimate the geometry of transparent objects, they fail to accurately model refraction effects and instead represent Fresnel effects as diffuse colors on the object's surface, as showcased in Fig. \ref{fig:NVS_refractive}.

We also conducted a benchmark of novel view images with backgrounds, utilizing the NeRF\cite{mildenhall2021nerf} and NeRO\cite{liu2023nero} as baselines for scenes with reflective objects, and NeRF and NeRFRO\cite{pan2022sampling} for scenes with refractive objects. Notably, we did not assess NeRO on the refractive objects since it is designed specifically for generating novel view images of reflective objects and lacks the capability to model refractive effects, as illustrated in Fig. \ref{fig:refr_failure_cases}(b). Similarly, NeRFRO\cite{pan2022sampling} was only evaluated on the reflective dataset for the same reason. The quantitative results are presented in Tab. \ref{tab:Quantitative_NVS}.

The quantitative results in Tab.~\ref{tab:Quantitative_NVS} demonstrate that our method excels in synthesizing novel view images when compared to the two baseline methods, both for reflective and refractive object datasets. These results highlight the superior performance of our approach in handling the Fresnel effect. Our evaluation reveals that while NeRO\cite{liu2023nero} exhibits a stronger ability to model the geometry of reflective cow, horse, and ball, it faces challenges when dealing with the geometry of the bunny, resulting in less accurate reconstruction quality in comparison to our method. This observation highlights the robustness of our approach, particularly in scenarios involving complex object shapes or the presence of implicit boundaries between objects and the background. For the refractive objects, our method also outperform the baseline methods both qualitatively and quantitatively, as illustrated in Fig. \ref{fig:NVS_refractive} and Tab. \ref{tab:Quantitative_NVS}.

\begin{table}
\caption{Quantitative results for novel view synthesis}
\centering
\scriptsize
\begin{tabular}{c|ccc|ccc}
    \toprule
     & \multicolumn{3}{c}{Reflective}
     & \multicolumn{3}{c}{Refractive} \\
    Method & PSNR$\uparrow$ & SSIM$\uparrow$ & LPIPS$\downarrow$ & PSNR$\uparrow$ & SSIM$\uparrow$ & LPIPS$\downarrow$  \\
    \midrule
    NeRF \cite{mildenhall2021nerf} & 23.536 & 0.685 & 0.237 & 26.183 & 0.764 &  0.217 \\
    NeRFRO \cite{pan2022sampling} & - & -&  - & 23.765 & 0.756 & 0.379 \\
    NeRO \cite{liu2023nero} & \textbf{30.839} & 0.805 & 0.196 & - & - & - \\
    NeRRF & 29.655 & \textbf{0.901} & \textbf{0.071} & \textbf{28.780} & \textbf{0.862} & \textbf{0.148} \\
   \bottomrule
\end{tabular}
\label{tab:Quantitative_NVS}
\vspace{1mm}

We compare our method with NeRF \cite{mildenhall2021nerf}, NeRO \cite{liu2023nero} and NeRFRO \cite{pan2022sampling}.
\end{table}

\subsection{Environment Radiance Estimation}

\begin{figure}[t]
\centering 
\includegraphics[width=1\linewidth]{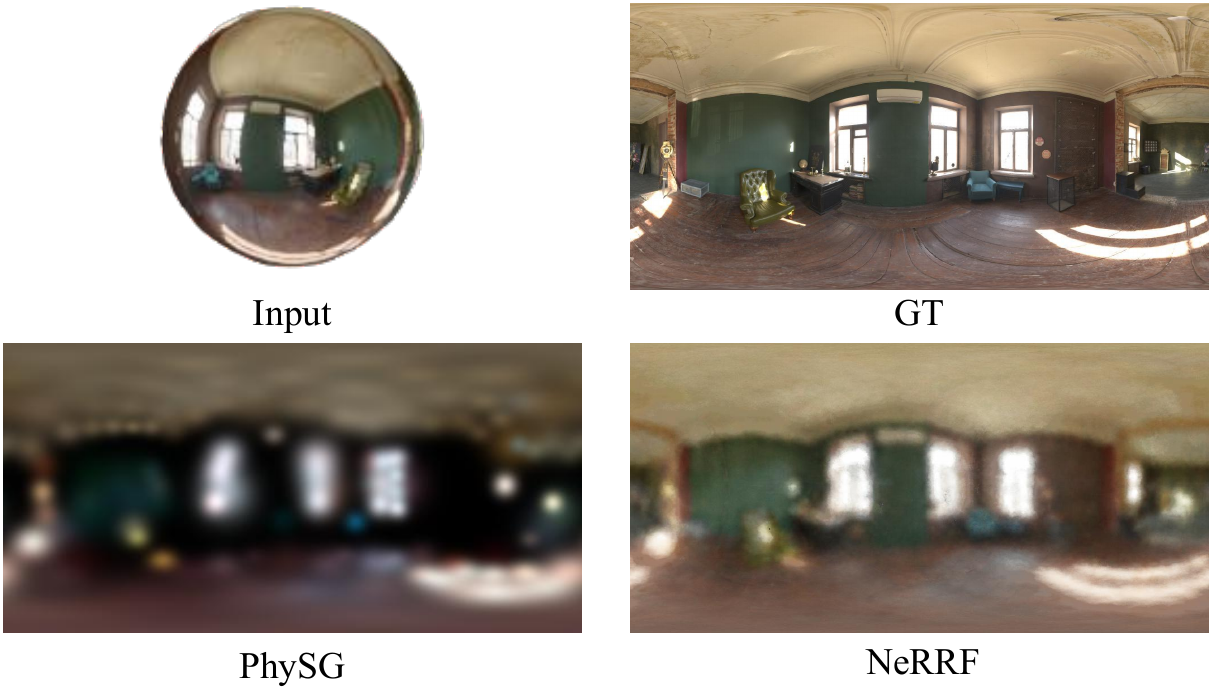}
\caption{Environment radiance estimation results. We generate environment map learned solely from the surface radiance of reflective ball. The result is compared with PhySG\cite{zhang2021physg}. }
\label{fig:envmap}
\end{figure}

Smooth glossy surfaces have the potential to function as visual sensors, enabling the observation of scenes out of individual's field of view. This has numerous practical applications in daily life, including tasks like estimating the environment behind a car through its rear-view mirror. Benefit from the decomposed geometry and radiance, NeRRF can effectively reconstruct scenes based on the outgoing radiance from the object surfaces. To validate this, we conduct an environment radiance estimation experiment, in which we mask out the background RGB information and learn the environment radiance solely from the object surface radiance. The input is illustrated in Fig. \ref{fig:envmap}. 

We compare NeRRF with PhySG, which can also learn the environment radiance from only the reflective surface, and the qualitative and quantitative results are shown in Fig. \ref{fig:envmap} and Tab. \ref{tab:envmap_estimation} respectively. The performance of NeRRF outperforms PhySG in all the metrics as shown in the table. Besides, the visualizations in Fig. \ref{fig:envmap} demonstrate that NeRRF can successfully estimate the environment map using only the reflection on the ball, while the result of PhySG is blurry. This is because our physically-based ray tracing module can accurately model the environment radiance, while PhySG approximates the environment map using Spherical Gaussian coefficients which are limited in frequency.

\begin{table}
\caption{Estimated environment radiance comparison}
\centering
\begin{tabular}{c|ccc}
    \toprule
    Method & PSNR$\uparrow$ & SSIM$\uparrow$ & LPIPS$\downarrow$ \\
    \midrule
    PhySG \cite{zhang2021physg} & 13.030 & 0.420 & 0.550 \\
    NeRRF & \textbf{16.468} & \textbf{0.479} & \textbf{0.433} \\
   \bottomrule
\end{tabular}
\label{tab:envmap_estimation}
\vspace{1mm}

We quantitatively compare the estimated environment radiance (learned solely by the object surface radiance) with the ground truth environment radiance.
\end{table}

\subsection{Scene Editing and Relighting}

\begin{figure*}[t]
\centering 
\includegraphics[width=0.7\linewidth]{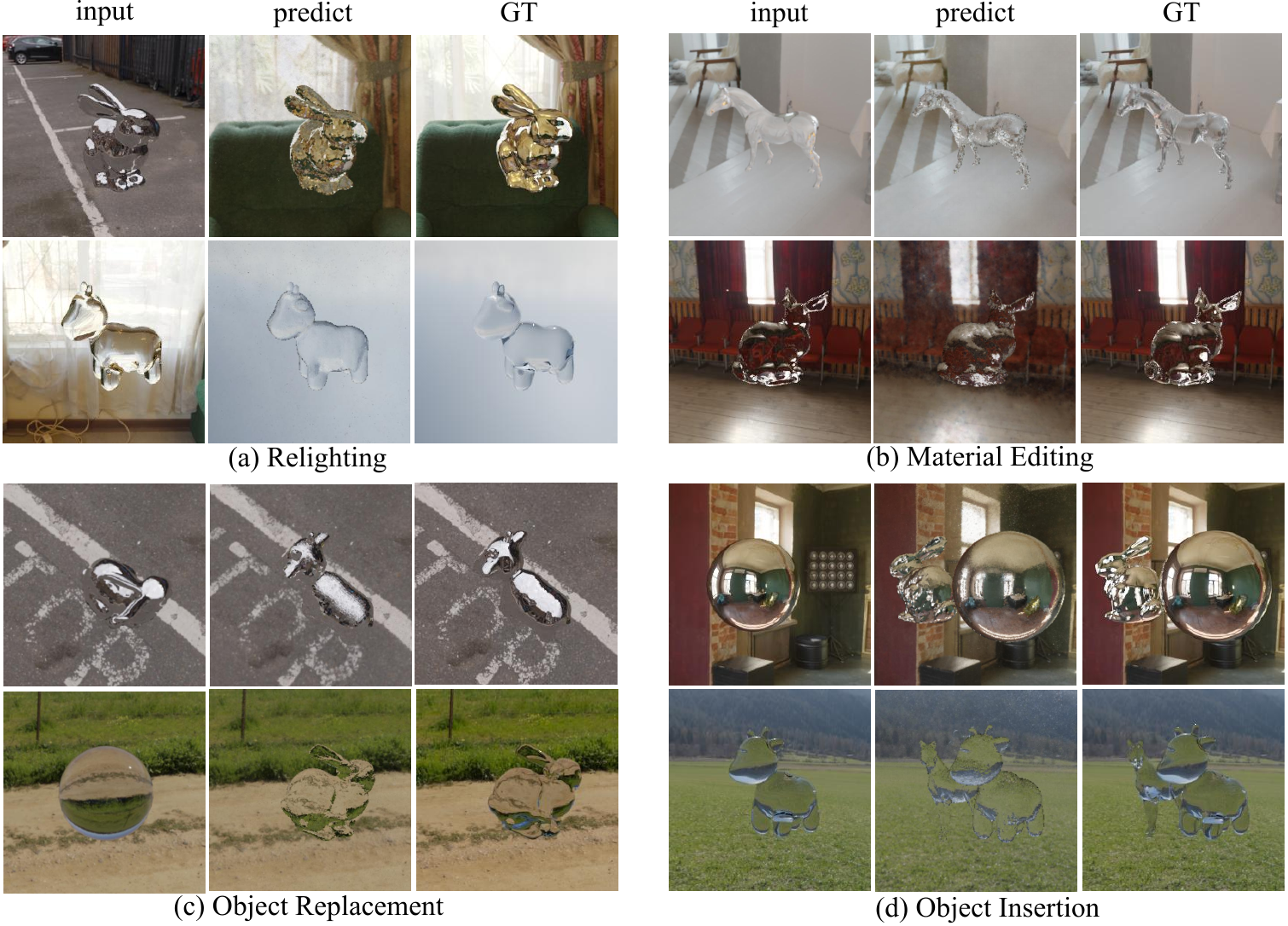}
\caption{Relighting and Editing Result. (a) Relighting We demonstrate the ability to change the background of a glossy bunny and a transparent cow. The new backgrounds are also rendered using NeRRF renderer. (b) Material Editing We showcase material editing by transforming a glossy horse into a transparent horse and changing the index of refraction (IOR) of a transparent bunny. These changes are achieved by editing the material properties in NeRRF. (c) Object Replacement We illustrate object replacement by transforming the glossy bunny into a glossy cow and the transparent ball into a transparent bunny, respectively. (d) Object Insertion We demonstrate object insertion by adding a bunny and a horse to two different scenes. }
\label{fig:relighting_editing}
\end{figure*}

Thanks to the disentanglement of object geometry and environment illumination in our approach, our method also possesses the capability to perform modifications on the input multi-view images. To demonstrate the practicability of our model, we conduct experiments on a set of image editing and relighting tasks. The qualitative results are presented in Fig. \ref{fig:relighting_editing}. Note that our rendered images are synthesized by the NeRRF renderer, while the reference images are the ones in the original scene which has the same camera calibration as the rendered images (not in the training set), and the ground-truth targets are generated using the Blender Cycles engine by directly editing the scene in Blender.

\subsubsection{Relighting} NeRRF enables controllable light by disentangling object material, geometry, and environment illumination. To validate this, we also demonstrate relighting results in Fig. \ref{fig:relighting_editing}(a). By changing the background and maintaining the optical properties of the objects, our method can directly render relighting results of transparent or glossy objects, which is achieved by changing the NeRRF model to the one trained with target lighting.

\subsubsection{Material editing}
Editing material properties of an object has applications for 3D shape design and visualization. We achieve material editing by straightforwardly changing the object material parameters involved in the ray-tracing process, and the results are provided in Fig. \ref{fig:relighting_editing} (b). In the first row, We edit the material of a horse from reflective to transparent, and in the second row, we change the object's index of refraction from 1.1 to 1.2. The generated image is basically consistent with the ground truth and conforms to the laws of optics, while the editing process is very simple.

\subsubsection{Object replacement} Object replacement is a critical application for high-quality augmented reality. Given a set of posed images of the scene containing a non-Lambertian object, we aim at replacing the object with a novel one, and rendering images with physically-correct light transport effects. To achieve this, we first learn the environment radiance using our model with the given set of images, and then replace the object geometry used in the ray-tracing process with the geometry of the novel object. Fig. \ref{fig:relighting_editing} (c) provides the qualitative results of object replacement, in the first row, we replace the metal bunny with a mental cow, and in the second row, we replace a transparent ball with a transparent bunny. As illustrated in the figure, our approach successfully models the non-Lambertian effects of the novel object and generates photorealistic renderings.

\subsubsection{Object insertion} Considering the time costs of the ray-tracing process, we conduct all the above experiments on scenes that consist of only one non-Lambertian object, but this doesn't mean our method cannot be applied to multi-object scenes. We carry out an experiment to insert a novel non-Lambertian object, which is achieved by providing the geometry and material of the new object for the ray-tracing module. As demonstrated in Fig. \ref{fig:relighting_editing} (d), we insert a reflective bunny into a scene with a reflective ball in the first row, while in the second row, we insert a transparent horse into the scene.  The results reveal that our model is capable of modeling multiple non-Lambertian objects with high fidelity, which can benefit a variety of VR/AR applications.

\subsection{Ablation on Network Architecture}

\subsubsection{Shape representation}\label{section: spherengp} 
\begin{figure}[t]
\centering 
\includegraphics[width=0.8\linewidth]{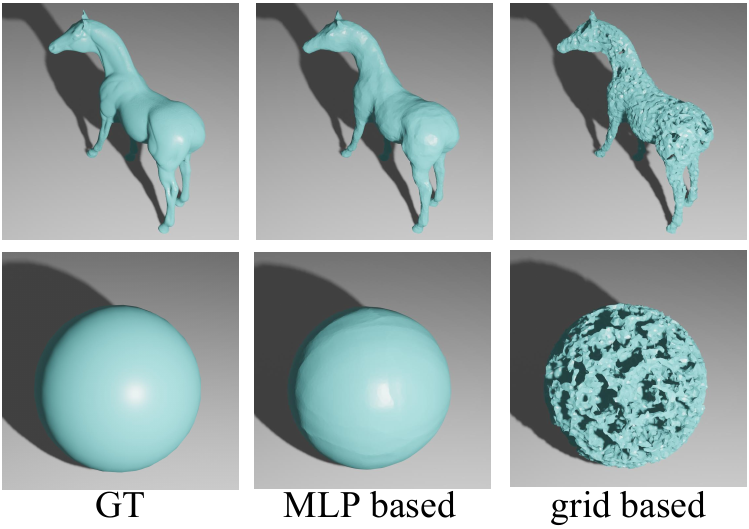}
\caption{Qualitative results on the ablation study of MLP-based DMTet. MLP-based method use a MLP to predict vertex SDF values and deformations $(s_i, \Delta v_i)$, while the grid-based method store $(s_i, \Delta v_i)$ in a vertex grid array.}
\label{fig:MLP_based_DMTet_ablation}
\end{figure}

With the Deep Marching Textrahedral(DMTet) as our shape representation, we embarked on two implementations: a MLP-based one and a grid-based one.
In the grid-based scheme, vertex SDF values and deformations $(s_i, \Delta v_i)$ utilized in \cite{shen2021dmtet} do not allow for the interpolation of SDF level sets from the four tetrahedron vertex SDF values and the calculation of SDF gradients, preventing the application of the eikonal loss. Hence, in the ablation study comparing the MLP-based representation with the grid representation, we opted not to apply the eikonal loss to the grid-based approach. All other settings remain the same. Tab. \ref{tab:MLP_based_DMTet_ablation} and Fig. \ref{fig:MLP_based_DMTet_ablation} presents the quantitative and qualitative results of the two implementations. As shown in Fig. \ref{fig:MLP_based_DMTet_ablation}, grid-based representation tends to generate surfaces riddled with holes, leading to a significantly higher chamfer distance compared to the MLP-based one. The presence of holes is also observed in the quantitative results for NDR\cite{munkberg2022extracting} of Fig. \ref{fig:geometry_comparison}. These holes persist even when relying on unreliable surface color supervision, indicating that the issue cannot be solely attributed to the inadequacy of color supervision. This underscores the critical role played by the MLP-based DMTet framework and the inclusion of the eikonal loss within our specific task domain.

\begin{table}
  \caption{Ablation study of MLP-based DMTet. }
  \centering
  \begin{tabular}{ccccc|c}
    \toprule
    Method & Bunny & Cow & Horse & Ball & Avg\\
    \midrule
    Grid-based & 0.0393 & 0.0341 & 0.0085 & 0.1611 & 0.0608 \\
    MLP-based & \textbf{0.0032} & \textbf{0.0024} & \textbf{0.0003} & \textbf{0.0037} & \textbf{0.0025} \\
    \bottomrule
  \end{tabular}
  \label{tab:MLP_based_DMTet_ablation}
\vspace{1mm}

We report the chamfer distance on ground truth mesh and extracted meshs as a quantitative measure for reconstructed geometry quality. The MLP-based method use a MLP to predict vertex SDF values and deformations $(s_i, \Delta v_i)$, while the grid-based method store them in a vertex grid array.
\end{table}

\subsubsection{Radiance representation} To evaluate the effectiveness of Sphere-NGP, we compare the rendering result on novel view synthesis using Sphere-NGP with two baselines: (1) NeRF-D: a variant NeRF with only direction as input and incorporates with our ray-tracing module; (2) NeRRF-P: a point-based reflective-refractive field with Instant-NGP as the backbone (not incorporates with the supersampling module due to memory efficiency). The experiment is conducted on two representative scenes: the reflective horse and the refractive cow, and the results are provided in \ref{tab:Quantitative_NVS}. As shown in the table, NeRRF outperforms NeRF-D with a significant margin, due to the multi-resolution hash encoding. It also ahicheves better performance compared with NeRRF-P. This is because point-based sampling is more difficult to converge, and the density learned is usually inaccurate, while our  direction-only formulation Eq. \ref{eq:env} is more compatible with the Blender dataset. Nonetheless, NeRRF-P is still effective in more complicated scenes where the environment cannot be seen as infinitely far, and our ray-tracing module serves as a plug-and-play module that can be applied to any NeRF-based method. 

\begin{table}
\caption{Ablation study of the Sphere-NGP}
\centering
\scriptsize
\begin{tabular}{c|ccc|ccc}
    \toprule
     & \multicolumn{3}{c}{Reflective Horse}
     & \multicolumn{3}{c}{Refractive Cow} \\
    Method & PSNR$\uparrow$ & SSIM$\uparrow$ & LPIPS$\downarrow$ & PSNR$\uparrow$ & SSIM$\uparrow$ & LPIPS$\downarrow$  \\
    \midrule
    NeRF-D & 32.619 & 0.931 & 0.077 & 24.574 & 0.780 & 0.185 \\
    NeRRF-P & 32.160 & 0.934 & 0.091 & 24.604 & 0.761 & 0.168 \\
    NeRRF & \textbf{35.103} & \textbf{0.954} & \textbf{0.068} & \textbf{25.465} & \textbf{0.798} & \textbf{0.181} \\
   \bottomrule
\end{tabular}
\label{tab:sngp_ablation}
\end{table}

\begin{figure}[t]
\centering 
\scriptsize
\includegraphics[width=0.8\linewidth]{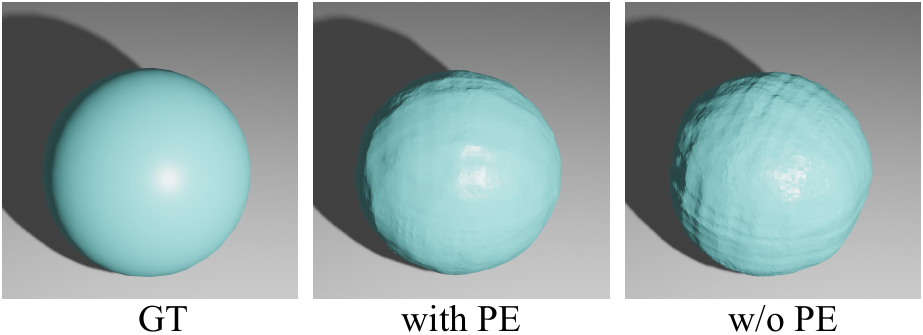}
\caption{Qualitative results on the effects of progressive encoding. 'PE' denotes the progressive encoding.}
\label{fig:progressive_encoding_ablation}
\end{figure}

\begin{table}
\caption{Ablation study of progressive encoding.}
\centering
\scriptsize
\begin{tabular}{ccccc|c}
\toprule
Method & Bunny & Cow & Horse & Ball & Avg\\
\midrule
w/o progressive encoding & 0.0323 & 0.0227 & 0.0094 & 0.0047 & 0.0172 \\
w/ progressive encoding & \textbf{0.0100} & \textbf{0.0213} & \textbf{0.0085} & \textbf{0.0040} & \textbf{0.0110} \\
\bottomrule
\end{tabular}
\label{tab:progressive_encoding_ablation}
\vspace{1mm}

We report the chamfer distance on ground truth mesh and extracted meshs as a quantitative measure for reconstructed geometry quality.
\end{table}

\subsubsection{Progressive encoding} \label{section:pe} We conduct an ablation study to investigate the role of progressive encoding. Because the encoding channels are fully unveiled after 500 iterations in our experiment setting, we train each object for 1000 iterations to assess the impact of progressive encoding. As the quantitative results presented in Tab. \ref{tab:progressive_encoding_ablation} indicate, the incorporation of progressive encoding leads to smaller chamfer distances between the predicted geometry and the ground truth geometry, demonstrating the effectiveness of progressive encoding. Additionally, visual inspection in Fig. \ref{fig:progressive_encoding_ablation} reveals that the sphere's surface appears smoother when progressive encoding is applied, in contrast to the version without progressive encoding. This indicates that progressive encoding facilitates the early learning of low-frequency features, which in turn aids in the removal of high-frequency noise.

\subsection{Ablation on Supersampling}

\begin{figure}[t]
\centering 
\includegraphics[width=0.8\linewidth]{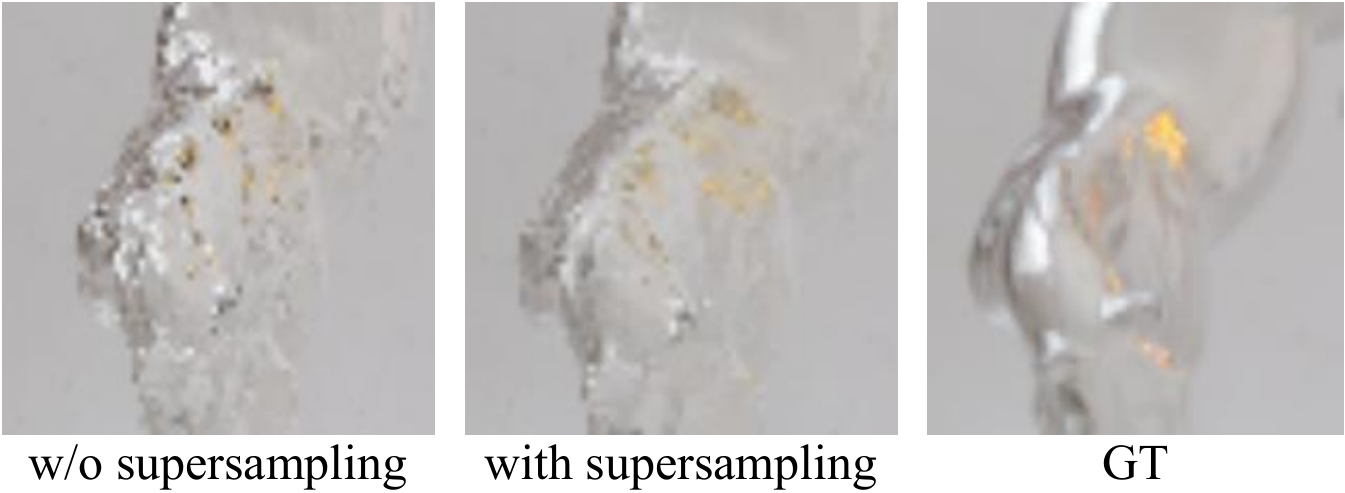}
\caption{Qualitative results on the effects of supersampling.}
\label{fig:supersampling_ablation}
\end{figure}

Tab. \ref{tab:supersampling_ablation} investigates the effects of our supersampling strategy. For comparison, we train a model without supersampling module which is denoted as ``w/o Supersampling".  As Tab. \ref{tab:supersampling_ablation} shows, the model with our supersampling strategy outperforms the "w/o Supersampling" in all scenes, which indicates that supersampling does help the learning of the scene representations. Fig. \ref{fig:supersampling_ablation} provides the qualitative results of supersampling. As shown in the figure, the rendering result of w/o supersampling is noisy on the horse's surface, which is caused by the unsmooth surface normal extracted from the predicted geometry. Instead, the result with supersampling becomes significantly smoother. This is because the radiance on the object's surface is the average radiance sampled in the virtual cone, which is equivalent to averaging the surface normal under a certain distribution. 

\begin{table}
\caption{Ablation study of supersampling.}
\centering
\scriptsize
\begin{tabular}{c|ccc|ccc}
\toprule
     & \multicolumn{3}{c}{\textbf{w/o} Supersampling}
     & \multicolumn{3}{c}{\textbf{w/} Supersampling} \\
    Object & PSNR$\uparrow$ & SSIM$\uparrow$ & LPIPS$\downarrow$ & PSNR$\uparrow$ & SSIM$\uparrow$ & LPIPS$\downarrow$  \\
\midrule
    Bunny-s & 25.738 & 0.874 & 0.076 & \textbf{26.583} & \textbf{0.878} & \textbf{0.064} \\
    Cow-s & 28.407 & 0.866 & 0.076 & \textbf{29.254} & \textbf{0.878} & \textbf{0.057} \\
    Horse-s & 33.325 & 0.949 & 0.079 & \textbf{35.102} & \textbf{0.954} & \textbf{0.068} \\
    Ball-s & 26.651 & 0.889 & 0.073 & \textbf{27.680} & \textbf{0.894} & \textbf{0.090} \\
\midrule
    Bunny-t & 28.132 & 0.859 & \textbf{0.076} & \textbf{28.606} & \textbf{0.861} & 0.142 \\
    Cow-t & 24.779 & 0.796 & \textbf{0.144} & \textbf{25.465} & \textbf{0.798} & 0.181 \\
    Horse-t & 30.169 & \textbf{0.920} & 0.185 & \textbf{30.397} & \textbf{0.920} & \textbf{0.183} \\
    Ball-t & 30.250 & 0.862 & 0.082 & \textbf{30.650} & \textbf{0.869} & \textbf{0.081} \\
\bottomrule
\end{tabular}
\label{tab:supersampling_ablation}
\end{table}

\begin{figure}[t]
\centering 
\includegraphics[width=0.5\linewidth]{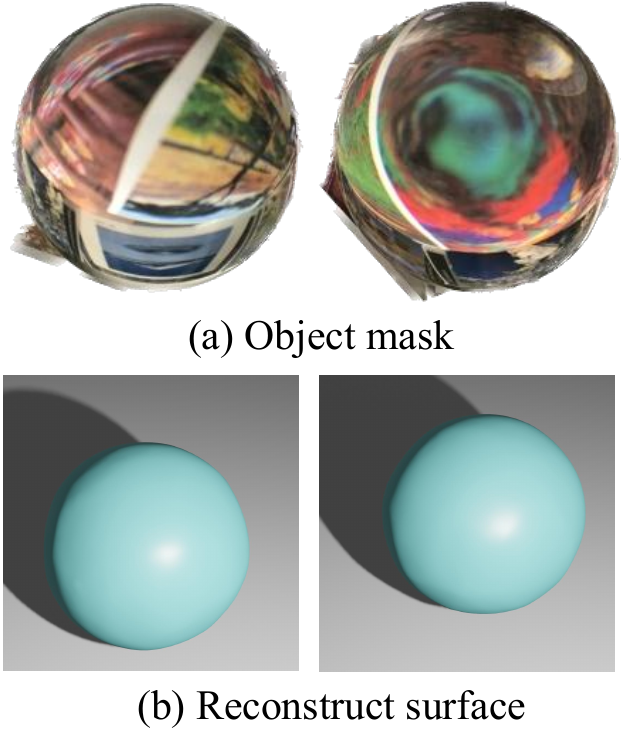}
\caption{Object mask and reconstructed surface. (a) Object mask obtained by the Apple's built-in cutout tool (b) Reconstructed surface which are not affected by the high-frequency noises of the object mask.}
\label{fig:real_geometry}
\end{figure}

\begin{figure}[t]
\centering 
\includegraphics[width=1\linewidth]{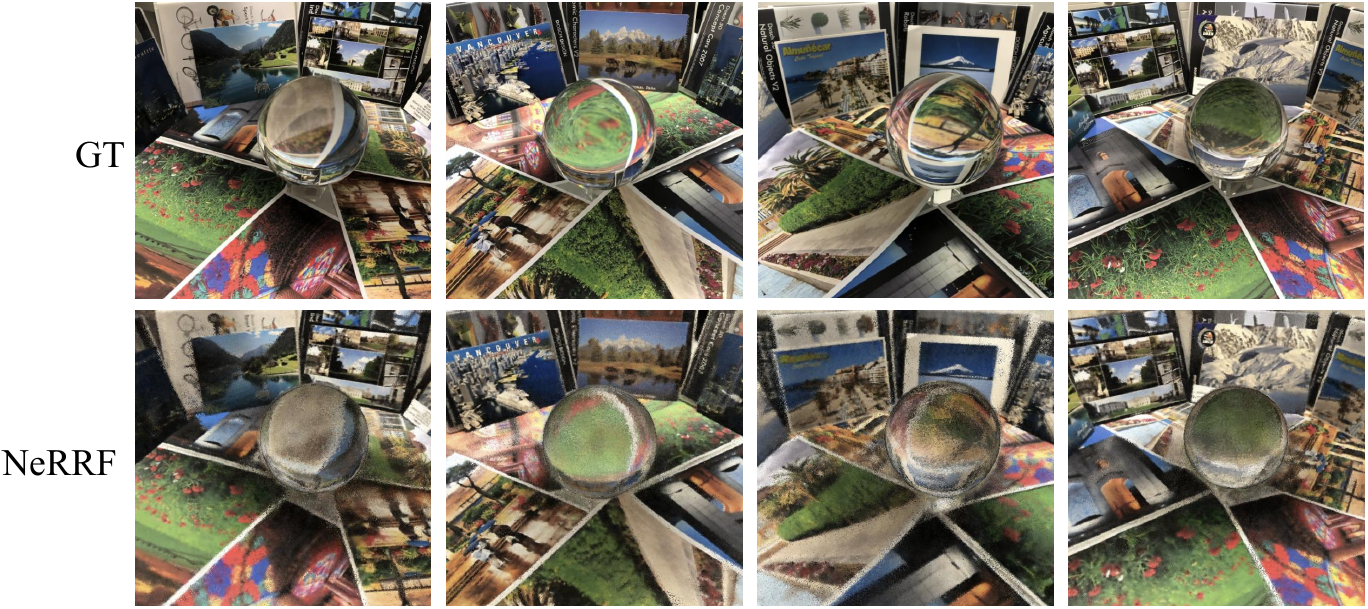}
\caption{Qualitative results of novel view synthesis on the real-world dataset provided by \cite{bemana2022eikonal}.}
\label{fig:NVS_real}
\end{figure}

\subsection{Real-world Application}

To further validate the effectiveness of NeRRF, we also applied it into a real-world dataset which is provided by \cite{bemana2022eikonal}.  Considering the environment illumination settings in real-world scenarios, we use NeRRF-P for the experiment. The qualitative novel view images are given in Fig. \ref{fig:NVS_real}. In this experiment, we access the object's masks through the Apple's built-in cutout tool \cite{applecutout}. While this tool may not provide precise masks for transparent objects, we observed that the predicted ball was unaffected by these artifacts. This demonstrates the robustness of our geometry reconstruction method, as illustrated in Fig. \ref{fig:real_geometry}.

\section{Conclusion}
In this paper, we  introduce NeRRF, a two-stage pipeline for geometry estimation and novel view synthesis of transparent and glossy objects. Through the use of a progressive encoder and eikonal loss applied to an MLP-based DMTet, our method can effectively reconstruct the geometry of non-Lambertian objects with only the supervision of object masks. NeRRF disentangles non-Lambertian object geometry from its highly view-dependent appearance, making it suitable for various tasks, including material editing, relighting, and environment illumination estimation, which could have implications for AR/VR applications. Our experiments have shown that NeRRF can successfully reconstruct object surfaces compared to state-of-the-art methods and generate high-quality novel view images. One limitation of our method is when learning geometry solely from the supervision of a mask loss may not be sufficient for accurately reconstructing some non-vertex surfaces. Additionally, NeRRF relies heavily on the availability of masks of non-Lambertian objects, which may not always be obtainable using current segmentation models.

\ifCLASSOPTIONcaptionsoff
  \newpage
\fi

{\small
\bibliographystyle{ieee_fullname}
\bibliography{egbib}
}





\end{document}